\documentclass{article}
\usepackage{ijcai19}

\usepackage{times}
\usepackage{soul}
\usepackage{url}
\usepackage[hidelinks]{hyperref}
\usepackage[utf8]{inputenc}
\usepackage[small]{caption}
\usepackage{graphicx}
\usepackage{amsmath}
\usepackage{booktabs}

\usepackage{amssymb}
\usepackage{pgfplots}
\usepackage{tikz}
\usepackage{verbatim}
\usetikzlibrary{shapes.gates.logic.US,trees,positioning,arrows}
\usepackage{algorithm}
\usepackage[noend]{algpseudocode}
\usepackage{cleveref}

\usepackage{bm}
\usepackage{amsbsy}

\newcommand{\argmax}{\operatornamewithlimits{arg\,max}}

\input{macros.tex}

\title{Monte-Carlo Tree Search for Policy Optimization}

\author{
Xiaobai Ma$^1$\and
Katherine Driggs-Campbell$^2$\and
Zongzhang Zhang$^3$\And
Mykel J. Kochenderfer$^1$\\
\affiliations
$^1$Aeronautics and Astronautics Department, Stanford University\\
$^2$Electrical and Computer Engineering Department, University of Illinois Urbana-Champaign\\
$^3$National Key Laboratory for Novel Software Technology, Nanjing University\\
\emails
maxiaoba@stanford.com,
krdc@illinois.edu,
zhangzongzhang@gmail.com,
mykel@stanford.edu
}

\begin{document}

\maketitle

\begin{abstract}
Gradient-based methods are often used for policy optimization in deep reinforcement learning, despite being vulnerable to local optima and saddle points. 
Although gradient-free methods (e.g., genetic algorithms or evolution strategies) help mitigate these issues, poor initialization and local optima are still concerns in highly nonconvex spaces.
This paper presents a method for policy optimization based on Monte-Carlo tree search and gradient-free optimization. 
Our method, called Monte-Carlo tree search for policy optimization (MCTSPO), provides a better exploration-exploitation trade-off through the use of the upper confidence bound heuristic.
We demonstrate improved performance on reinforcement learning tasks with deceptive or sparse reward functions compared to popular gradient-based and deep genetic algorithm baselines.
\end{abstract}

\section{Introduction}

Recent advances in deep reinforcement learning (DRL) have shown success on a variety of tasks, including games~\cite{DQN} and robotics~\cite{ACKTR}.
Artificial neural networks (ANNs) are often used to represent either a state-action value function~\cite{rainbow} or a policy network~\cite{TRPO,PPO}, 
and they are typically optimized through gradient-based methods~\cite{adam}. 
While gradient-based optimization can be extremely efficient on tasks with smooth or incremental rewards, they often require reward shaping~\cite{reward_shaping}.
Such solvers struggle in environments with sparse or deceptive rewards, indicating that issues such as local optima and saddle points are limiting their success~\cite{GA}. 

Recently, a category of gradient-free algorithms produced policies that outperformed those trained using the typical gradient-based methods on certain tasks~\cite{es_benchmark}.
The success of these methods implied that gradient-free algorithms could also work on deep and complex neural networks.
Evolution Strategies (ES) were used to evolve a policy network by adding noise to the neural network parameters and performing parallel rollouts~\cite{ES}. The evolution direction, i.e., the parameter update vector, is determined using the accumulated rewards of perturbed trajectories. Such strategies are tolerant to delayed rewards and long horizons because they directly use the trajectory return, in contrast that gradient-based methods often need immediate rewards for better credit assignment.

ES is similar to a gradient approach using finite differences over trajectory returns. As a truly gradient-free method, the deep genetic algorithm (Deep GA) applies a simple genetic algorithm to evolve neural networks~\cite{GA}. Deep GA maintains a population of neural network parameters. The top performing individuals are iteratively mutated by adding random noise to create the next generation. 

The success of Deep GA on difficult DRL tasks suggests that, in some cases, following the gradient direction may be misleading, especially in highly nonconvex spaces.
Other population-based methods incorporate and adapt gradient-based approaches in the mutation steps~\cite{ERL}.
While this combined approach improves data efficiency, it also loses some randomness in the mutation and, more importantly, lacks the scalability to large population sizes and neural networks with deep structure.

This paper formulates the policy optimization problem as a deterministic Markov decision process (MDP) and uses Monte-Carlo tree search (MCTS) to find the optimal trajectories in the policy space. 
Using MCTS on population based methods has been applied to games with discrete state and action spaces~\cite{EMCTS}, but has not been effectively generalized to high-dimensional continuous state and action spaces.

Using a similar mutation process and network parameter representation as \citet{GA}, the proposed method, referred to as Monte-Carlo tree search for policy optimization (MCTSPO), is shown to be a more efficient tree search algorithm than Deep GA. 
MCTSPO improves the exploration-exploitation tradeoff by adopting the UCT principle in choosing parents~\cite{UCT}, i.e., the individual with the highest upper confidence bound gets expanded first. 

MCTSPO is compared to a state-of-the-art gradient based algorithm, trust region policy optimization (TRPO) \cite{TRPO}, and a genetic algorithm with safe mutation \cite{safemutation} on multiple continuous control tasks. 
We demonstrate how our method increasingly outperforms these baselines as the difficulty of the tasks increases.

In summary, we present the following contributions:
\begin{enumerate}
    \item We format the policy optimization task as an MDP, which enables the application of existing MDP solvers to policy optimization;
    \item We introduce new modifications to the MCTS algorithm to increase its efficiency and scalability when solving the policy optimization MDP; and
    \item We show that MCTSPO is able to outperform existing methods on a set of difficult reinforcement learning benchmark tasks.
\end{enumerate}

This paper is organized as follows: \Cref{sec:background} gives an introduction to MDPs and Deep GA. 
\Cref{sec:methods} describes how to formulate policy optimization as an MDP and presents MCTSPO. 
\Cref{sec:experiments} introduces the experimental setup and implementation details. The results and the corresponding discussion are presented in \Cref{sec:results}. 
\Cref{sec:conclusion} summarizes the contributions of the work and future work. 

\section{Background}
\label{sec:background}

\subsection{Markov Decision Process}
A Markov decision process (MDP) is a common model for sequential decision making problems~\cite{MDP}. 
An MDP is defined as a tuple $M=(S,A,P,R,T,\gamma,\rho_0)$, where $S$ is the state space; $A$ is the action space; $P: S\times A \times S \rightarrow [0,1]$ defines the transition probability; $R: S\times A \times S \rightarrow \mathbb{R}$ is the reward function; $T$ is the horizon; $\gamma \in (0,1]$ is the discount factor; and $\rho_0: S \rightarrow [0,1]$ is the initial state distribution. 
A trajectory $\tau$ is a tuple of sequential transitions in the MDP: $\tau=((s_0,a_0,r_0,s_1),(s_1,a_1,r_1,s_2),\ldots,(s_{T-1},a_{T-1},r_{T-1},s_T))$. 
The return of the trajectory is given by $\rho(\tau)=\sum_{t=0}^{T-1}\gamma^tr_t$.

A policy $\pi:S \times A \rightarrow [0,1]$ is a mapping from each state in $S$ to a probability distribution over the actions. 
Let $\eta(\pi)$ denote its expected discounted reward:
\begin{equation}
\label{eq:policy return}
\eta(\pi)=\mathbb{E}_{s_0,a_0,\ldots}\Big[\sum_{t=0}^{T-1} \gamma^t R(s_t,a_t,s_{t+1})\Big],
\end{equation}
where $s_0 \sim \rho_0(s), a_t \sim \pi(a_t \mid s_t)$, and $s_{t+1} \sim P(s_{t+1} \mid s_t,a_t)$.

\subsection{Deep Genetic Algorithm}

Deep GA evolves a population of neural network policies, which are candidate optimal policies. 
At each iteration, each individual policy in the population is deployed in the environment to receive a fitness score, i.e., the average return of the rollout policy.
Then, Deep GA performs truncation selection to choose the top $k$ individuals as the parents of the next generation. 
The individual in the next generation is generated by uniformly choosing one parent and performing a mutation. 

The mutations are in the form of additive Gaussian noise on the parent's parameter vector: $\theta'=\theta + \sigma\epsilon$, where $\theta$ and $\theta'$ are parameter vectors of the parent and the child's neural network respectively, $\sigma$ is the step size, and $\epsilon$ is sampled from a standard multivariate Gaussian distribution. 
Typically, a crossover step is also performed before mutation to combine the parameters of two parents to generate an offspring. Deep GA skips this step, since there is no trivial way to efficiently combine the learned information of two neural networks. Due to the high non-linearity of the neural networks, switching parameters arbitrarily between two neural networks could destruct the encoded information from both sides.

In a naive implementation, this method would not scale to large populations. Instead of explicitly storing parameter vectors of the entire population, Deep GA stores the series of random seeds that initialize and mutate the network. This idea is borrowed in MCTSPO and explained in \Cref{sec:methods}.

\section{Methods}
\label{sec:methods}

\subsection{Tree Structure Formulation}

\begin{figure}
    \centering
    \scalebox{0.7}{
\begin{tikzpicture}[
    ]
   \begin{scope}[xshift=-7.5cm,yshift=-5cm,very thick,
		node distance=1.6cm,on grid,>=stealth',
		comment/.style={text,draw},
		comp/.style={circle,draw,fill=blue!40},
		comp2/.style={circle,draw,fill=orange!40},
		comp3/.style={circle,draw,fill=red!40}]
   \node [comp2]	 (root)	{};  
   \node [comp3]	 (g11)	[below=of root,xshift=-3.2cm]	{} edge [<-] (root);
   \node [comp]	 (g12)	[right=of g11]			{} edge [<-] (root);
   \node [comp3]	 (g13)	[right=of g12]			{} edge [<-] (root);
   \node [comp]	 (g14)	[right=of g13]			{} edge [<-] (root);
   \node [comp3]	 (g15)	[right=of g14]			{} edge [<-] (root);
   \node [comp]	 (g21)	[below=of g13,xshift=-3.2cm]	{} edge [<-] (g11);
   \node [comp]	 (g22)	[right=of g21]			{} edge [<-] (g11);
   \node [comp]	 (g23)	[right=of g22]			{} edge [<-] (g13);
   \node [comp]	 (g24)	[right=of g23]			{} edge [<-] (g13);
   \node [comp]	 (g25)	[right=of g24]			{} edge [<-] (g15);
    \node[text width=0.8cm] [left=of root,xshift=-3.7cm] {\huge Root};
    \node[text width=3.2cm] [left=of g13,xshift=-4cm] {\huge Generation (depth) 1};
    \node[text width=3.2cm] [left=of g23,xshift=-4cm] {\huge Generation (depth) 2};
   \end{scope}
\end{tikzpicture}}
    \caption{\small Tree structure in Deep GA. Each node represents an individual, and each arrow represents a mutation. Pink nodes are selected as parents for the next generation.}
    \label{fig:ga_tree}
\end{figure}
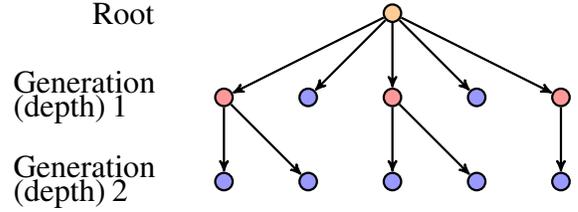

Since there is no crossover in Deep GA, each individual has a single parent from the previous generation, and may have several children in the next generation. 
If we add a dummy common parent to all individuals in the first generation, then we may represent the population evolution as a tree. 
In the tree, all individuals are represented by nodes and each edge represents a mutation. 
Each generation corresponds to a tree depth. 
Figure~\ref{fig:ga_tree} shows the tree structure formed by a genetic algorithm with population size 5 and truncation size 3. 

We observe that the genetic algorithm follows a simple tree search heuristic. 
At each depth, only a subset of nodes (the nodes selected as elites) is expanded and the rest of them are discarded. 
The number of nodes at each depth is fixed, determined by the population size. 
We introduce the idea of using Monte-Carlo tree search (MCTS) in the policy space, which balances exploration and exploitation using the upper confidence bound heuristic~\cite{UCT}.

\subsection{Policy Optimization MDP}
Formally, we consider the parameter optimization of a policy network as an MDP. 
There are two MDPs under consideration: one defined by the task environment (task MDP), and the other defined for policy optimization (policy optimization MDP). To differentiate, we use a tilde to denote elements related to the task MDP. Unless specified, all MDPs mentioned in the following text refer to the policy optimization MDP.

The state space of the MDP is the parameter space of the neural network. Each state is represented as the parameter vector of the network.
The action of the MDP is the mutation applied to the network. 
Taking inspiration from \citet{GA} and \citet{safemutation}, each action is represented as a vector consisting of a random seed and a magnitude, i.e., $a=[\text{seed},v]^\top$.

The transition of the MDP is deterministic and is performed as follows.
To create the first generation from the root, we use the seed to initialize the neural network parameters with preferred weight initialization functions.
For later generations, we first use the seed to randomly generate a direction vector $\hat{\delta}$ in the parameter space using the standard multivariate normal distribution.
Then, the new state is given by $s_{t+1}=s_{t}+v\hat{\delta}$, where $v$ is the mutation magnitude specified by the action.
Thus, we may reproduce the parameter vector for any node in the tree by performing mutations specified by the action sequence connecting the root and the node. 
This representation avoids explicitly saving all nodes' parameters and makes the algorithm tractable. 

The reward is defined as each mutation's performance improvement. For the tuple $(s,a,s')$, $R(s,a,s')=\tilde{\eta}(s')-\tilde{\eta}(s)$, where $\tilde{\eta}(s)$ is the expected return in the task environment produced by the policy specified by $s$. 
We assume $\tilde{\eta}(s_0)=0$ for the root state. Assuming no discount, $\gamma=1$, the return of a trajectory $\rho(\tau)$ equals the expected return of the last state, $\tilde{\eta}(s_T)$. Thus, we only need to rollout the policy represented by $s_T$ in the task environment to get the trajectory return.

\subsection{MCTS Methodology}
We use MCTS to solve the MDP~\cite{UCT}. The original algorithm is omitted here. This section focuses on introducing the important modifications to make MCTS work for policy optimization.

Since the action space is continuous, we use \textit{progressive widening} to control the tree expansion~\cite{progressive_widening}. We do not need to constrain the state widening since the state transition is deterministic. Similar to \citet{MCTSdpw}, we constrain the number of actions at each state node with $|N(s,a)|<kN(s)^\alpha$, where $N(s)$ is the number of times that $s$ has been visited; $N(s,a)$ is the number of times that $a$ has been chosen as the next action at state $s$; and $|N(s,a)|$ is the number of different actions tried at state $s$. Both positive parameters $k$ and $\alpha$ are used to control the widening of the tree.

MCTS requires two meta policies: one for action selection and one for rollout. 
For action selection, we follow the upper confidence bound principle: 
\begin{equation}
\label{eq:UCB}
a'\gets \argmax_a Q(s,a)+c\sqrt{\frac{\log N(s)}{N(s,a)}},
\end{equation}
where $c$ is an exploration parameter.  The estimated return of choosing $a$ at state $s$ (in no discount and deterministic MDP) is $Q(s,a)=R(s,a,s')+V(s')$, where $V(s')$ is the estimated return at next state $s'$. 

In the original MCTS algorithm, $V(s')$ is estimated by the \textit{average} return collected by the subtree rooted at $s'$. 
Here, we use a variant of the MaxUCT estimation in the deterministic case, where $V(s')$ is the \textit{maximum} return collected by the subtree~\cite{MaxUCT}. 
This estimation improves performance when the variance of returns among siblings may be very large.
This large variance is common in policy space due to small differences and errors compounding during rollout, leading to diverging trajectories and returns.
 
For the rollout policy, a traditional approach is to apply random mutations to the current policy, and use the return of the final trajectory as an estimate of the value of the current policy, $V(s)$. This estimate is unbiased but has large variance. Thus, instead of random mutations, we employ a \emph{no mutation} strategy. That is, in rollout, we just deploy the current policy in the environment and return the rewards. This approach requires less computation, and has lower variance.

Finally, we apply a technique called \textit{safe mutation} to constrain the average performance divergence of a single mutation~\cite{safemutation}. 
Due to the nonlinear nature of neural networks, adding random noise to neural network parameters can significantly and unpredictably influence its output. 
Constraining the average performance divergence ensures that each branch in the tree corresponds to a local region of the policy space, instead of distributing in the entire policy space randomly.

The performance divergence is defined as the expected mean square difference between the new and original policy outputs~\cite{safemutation}. It is estimated by:
\begin{equation}
    D(s,s_{old})=\frac{1}{\tilde{T}}\sum_{t=1}^{\tilde{T}} \sum_{k=1}^{|\tilde{A}|} \Big[NN(\tilde{s_t};s)_k-NN(\tilde{s_t};s_{old})_k\Big]^2,
\end{equation}
where $s$ and $s_{old}$ are the new and the original policy parameters. Recall $s=s_{old}+v\hat{\delta}$, where $v$ and $\hat{\delta}$ are the mutation magnitude and direction. $NN(\tilde{s_t};s)_k$ is the $k$-th output of the policy network parameterized by $s$ given input $\tilde{s_t}$. $\tilde{T}$ is the task environment's time horizon. $|\tilde{A}|$ is the dimension of the task environment's action space; and $\tilde{s_t}$, where $t\in \{1, 2, \ldots, \tilde{T}\}$ is the sampled state at step $t$ using the current policy. 


Given $\hat{\delta}$, $s$, and the divergence constraint $D_{max}$, we solve the mutation magnitude efficiently using methods similar as in~\citet{TRPO}. The divergence $D$ is approximated by $\hat{D}$ with:
\begin{equation}
    \hat{D}(s,s_{old})=\frac{1}{2}(s-s_{old})^\top U(s-s_{old}),
\end{equation} 
where $U_{ij}=\frac{\partial}{\partial s_i}\frac{\partial}{\partial s_j}D(s,s_{old})|_{s=s_{old}}$. Then $v=\sqrt{\frac{2D_{max}}{\hat{\delta}^\top U \hat{\delta}}}$, where $\hat{\delta}^\top U \hat{\delta}$ can be calculated efficiently as introduced in \citet{TRPO}. 
Then, a line search is performed on $v$ to ensure the satisfaction of the performance divergence constraint.

Calculating divergence needs rollouts in the task environment.
Therefore, when expanding a node, we either must resample the trajectory or store the trajectory for each node when first reached. 
The first way is not sample efficient, since we deploy duplicate policies in the environment.
The second suggestion, however, is inefficient in terms of memory.

To overcome this, we calculate feasible candidate actions when we first reach the node. 
When we want to expand a node, we randomly choose one from its candidate actions. 
Thus, we only need to store candidate actions for each node, which requires much less memory than storing the trajectory while avoiding duplicated sampling. For a node $s$, we add an additional buffer $CA(s)$ that stores the candidate actions. When a new state $s$ is added to the tree, rollout is performed on $s$ and a trajectory $\tau_s$ is collected from the task environment. We use $\tau_s$ to calculate $n_\text{ca}$ candidate actions and store them in $CA(s)$, where $n_\text{ca}$ is a predetermined constant. When we want to expand $s$ with new actions, we simply pop an action from $CA(s)$. When $CA(s)$ is empty, we need to recollect a trajectory from $s$ and calculate a new set of $CA(s)$. Compared to sampling trajectories every time when expanding a state node, this increases the sample efficiency up to $n_\text{ca}$ times.

In summary, based on the original MCTS, we apply progress widening on the actions. We use MaxUCT for value estimation. We deploy \textit{no mutation} for rollout. We add performance divergence constraints on the mutation magnitude and use precalculated candidate actions to improve sample and memory efficiency.
\Cref{algo:mctspo} outlines the learning procedure for MCTSPO. \Cref{algo:rollout,algo:getca} show how to rollout and calculate candidate actions. 

\begin{algorithm}[t!]
\caption{MCTS for Policy Optimization (MCTSPO)}
\label{algo:mctspo}
\begin{algorithmic}

\Function{MCTSPO}{Task environment $\Gamma$, Initial state $s_0$}
    \State $Tree \gets \emptyset$
    \For{$i \gets 1$ \textbf{to} $n_\text{itr}$}
        \State \Call{Simulate}{$Tree$, $\Gamma$, $s_0$}
    \EndFor
    \State \Return{$s^*\gets \argmax_{s\in Tree}\tilde{\rho}(\tau_s)$}
\EndFunction
   
\Function{Simulate}{$Tree$, $\Gamma$, $s$}
    \If{$s \notin Tree$}
        \State $Tree \gets Tree \cup \{s\}$
        \State $(N(s),A(s),CA(s)) \gets (0,\emptyset,\emptyset)$
        \State \Return{\Call{Rollout}{$\Gamma$, $s$}}
    \EndIf
    \State $N(s) \gets N(s)+1$
    \If{$|N(s,a)|<kN(s)^\alpha$}
        \If{$CA(s) = \emptyset$}
            \State $\tau_s \gets \Call{Sample}{\Gamma, s}$
            \State $CA(s) \gets \Call{GetCA}{s, \tau_s}$
        \EndIf
        \State $a \gets \Call{Pop}{CA(s)}$
        \State $(N(s,a),Q(s,a)) \gets (0,Q_0(s,a))$
        \State $A(s) \gets A(s) \cup \{a\}$
    \EndIf
    \State $a' \gets \argmax_{a\in A(s)} Q(s,a)+c\sqrt{\frac{\log N(s)}{N(s,a)}}$
    \State $N(s,a') \gets N(s,a')+1$
    \State $s' \gets \Call{Mutate}{s,a'}$
    \State $q \gets \Call{Simulate}{Tree,\Gamma,s'}$
    \If{$q > Q(s,a')$}
        \State $Q(s,a') \gets q$
    \EndIf
    \State \Return{$Q(s,a')$}
\EndFunction
\end{algorithmic}
\end{algorithm}

\begin{algorithm}[t!]
\caption{Rollout}
\label{algo:rollout}
\begin{algorithmic}
\Function{Rollout}{$\Gamma$, $s$}
    \State $\tau_s \gets \Call{Sample}{\Gamma, s}$
    \State $CA(s) \gets \Call{GetCA}{s, \tau_s}$
    \State \Return{$\tilde{\rho}(\tau_s)$}
\EndFunction
\end{algorithmic}
\end{algorithm}

\begin{algorithm}[t!]
\caption{Get candidate actions}
\label{algo:getca}
\begin{algorithmic}
\Function{GetCA}{$s$, $\tau_s$}
    \State $CA \gets \emptyset$
    \For{$i \gets 1$ \textbf{to} $n_\text{ca}$}
        \State $seed \gets \Call{RandSeed}{}$
        \State $v \gets \Call{GetMagnitude}{s,\tau_s,seed}$
        \State $CA \gets CA \cup \{[seed,v]^{\top}\}$
    \EndFor
    \State \Return{CA}
\EndFunction
\end{algorithmic}
\end{algorithm}

\section{Experiments}
\label{sec:experiments}
We compare the performance of MCTSPO to two state-of-the-art baselines: TRPO as a representative of gradient-based methods~\cite{TRPO} and Deep GA using \textit{safe mutation}~\cite{safemutation}. The test environments are divided into two classes: classic control and Roboschool.

\subsection{Classic Control}
Three classic continuous control tasks, Acrobot~\cite{Acrobot}, Mountain Car~\cite{mountaincar}, and Bipedal Walker~\cite{openaigym}, are tested with the same time horizon $\tilde{T}=100$ and reward function. The agent receives a positive reward of 1.0 if the goal is reached, otherwise a small control penalty proportional to the action magnitude is applied.
With this reward function, there is a trivial local optimum where zero control effort is applied at each step with zero reward. 
Some of the task environments' parameters are changed to increase the difficulty of the task:
\begin{itemize}
    \item[-] Acrobot: We increase the vertical position that needs to be reached for the end point of the robot arm, $y_\text{goal}$. The maximum reachable height is 2, meaning the two arms need to be strictly vertical. Here we set $y_\text{goal}=1.999$.
    \item[-] Mountain Car: We decrease the power of the car, $p_\text{car}$. A smaller $p_\text{car}$ means a smaller acceleration limit in one time step, which makes reaching the flag harder. Here $p_\text{car}$ is set to \num{0.0015}.
    \item[-] Bipedal Walker: We increase the horizontal length that the robot needs to move forward for success, $x_\text{goal}$. $x_\text{goal}$ is set to 30 in our experiments.
\end{itemize}



\subsection{Roboschool Robotics}
To test the performance on high dimensional tasks, we adapt three robotics environments, Ant~\cite{Ant}, HalfCheetah~\cite{HalfCheetah}, and Hopper~\cite{Hopper}, from OpenAI Roboschool~\cite{PPO}. 
The environment goal is to control the robot to walk forward for \SI{2}{\meter}.
The agent receives a positive reward of 1000 when the goal is achieved, otherwise it is penalized for control effort and invalid actions like feet collisions. 
The time horizon $\tilde{T}$ is set to 500.

\subsection{Algorithm Implementation}
For TRPO, we use the Gaussian multilayer perceptron architecture from RLLab~\cite{rllab} with hidden layer sizes of 128, 64, and 32 with $\tanh$ activations. It is trained for 5000 iterations using step sizes 0.1 and 1.0. The batch size is set to 1000 for classic control tasks and 5000 for Roboschool.

For Deep GA, we use the deterministic multilayer perceptron architecture with the same network structure as used in TRPO. The population sizes are 100, 500, and 1000 with 500, 100, and 50 training iterations, respectively. The truncation size for parent selection is 20. At each iteration, the top three individuals persist to the next generation with no mutation, following a technique called \textit{elitism}~\cite{GA}. Since both the environment and the policy are deterministic, we use a batch size which is equal to the horizon, i.e., only one trajectory is sampled per rollout. The divergence constraint for the mutation step is set to 1.0 through preliminary tests. 

For MCTSPO, we use the same architecture and batch size as in Deep GA. We use an exploration constant of $\sqrt{2}$ for classic control tasks and 10 for Roboschool.\footnote{In the case of an MDP with finite state and action spaces and returns in the range of $[0,1]$, an exploration constant of $\sqrt{2}$ gives a theoretical guarantee of finding the (global) optimal policy~\cite{UCT}. In classic control, we consider environments with a continuous state and action space and the returns are approximately in this range.  Nonetheless, we observed reasonable results with this parameter. For Roboschool robotics, the return ranges are much larger, so we scaled the exploration constant accordingly.}  
The progressive widening parameters are set to $\alpha=k=0.3$, 0.5, and 0.8, respectively.\footnote{Note that $\alpha$ and $k$ are not necessarily equal.}
The number of candidate actions is set to $n_\text{ca}=4$ to balance the computation complexity and the sample efficiency. We train for 50,000 iterations with the same divergence constraint as used in Deep GA.

All three algorithms have the same number of environment calls in total for the training.\footnote{Although the total number of environment calls is the same for each algorithm, the training clock time for Deep GA and MCTSPO is approximately twice that of TRPO. This difference is mainly caused by the single-threaded sampling in our Deep GA and MCTSPO implementation. Parallel MCTS is an important area of future work.}

\section{Results}
\label{sec:results}
\begin{figure*}[!t]
    \centering

    \includegraphics[width = .665\columnwidth]{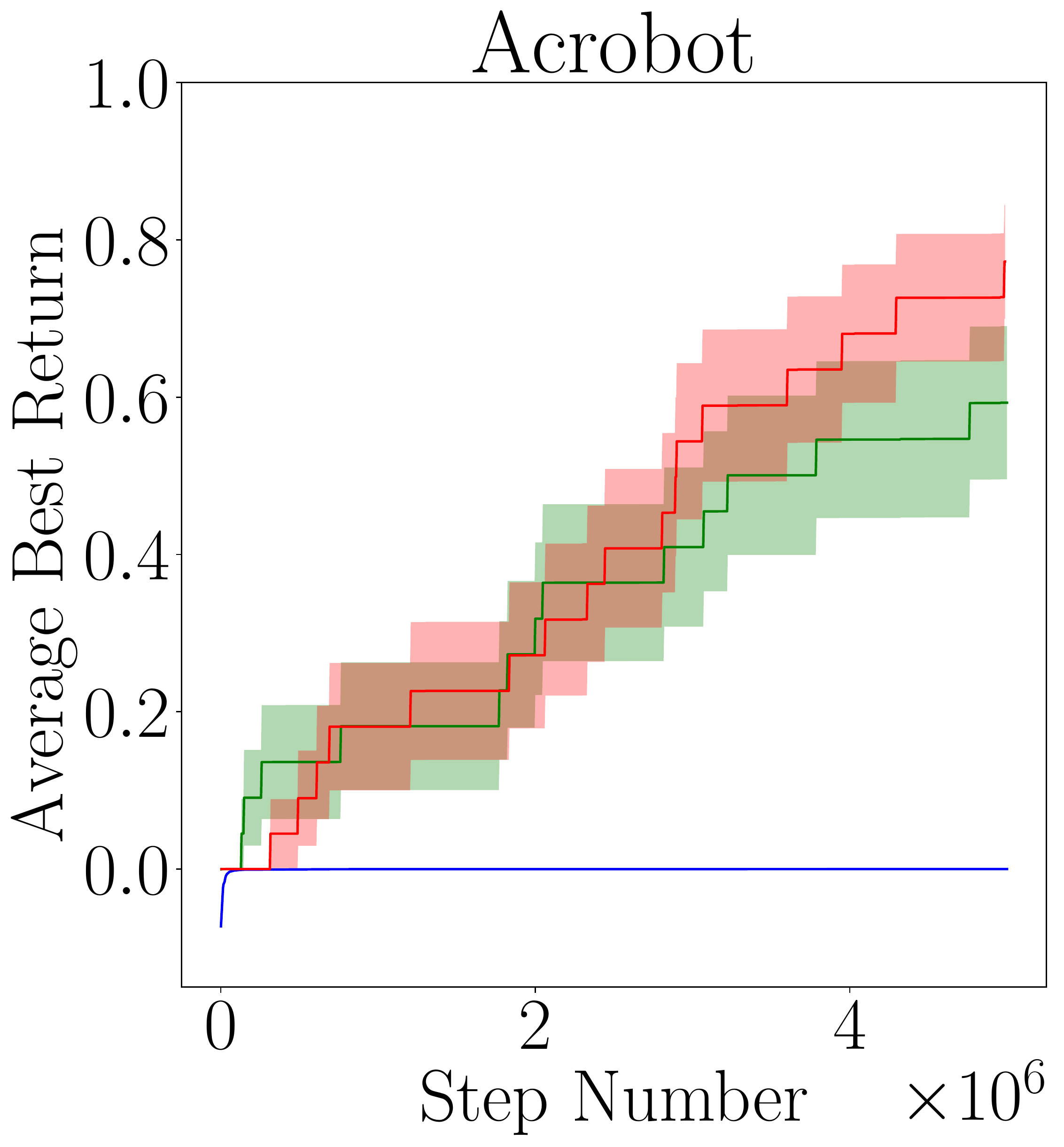}
    \includegraphics[width = .665\columnwidth]{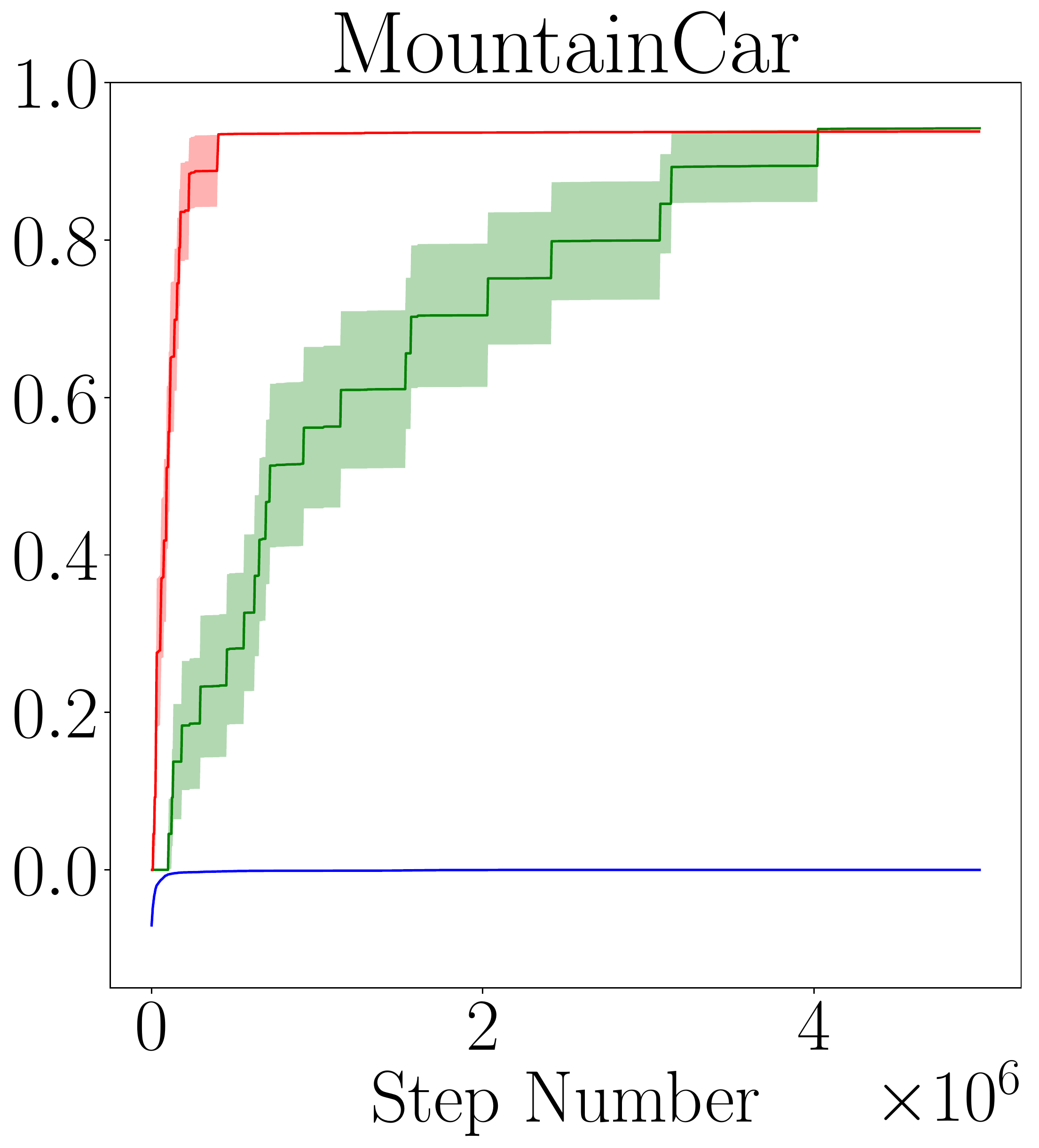}
    \includegraphics[width = .665\columnwidth]{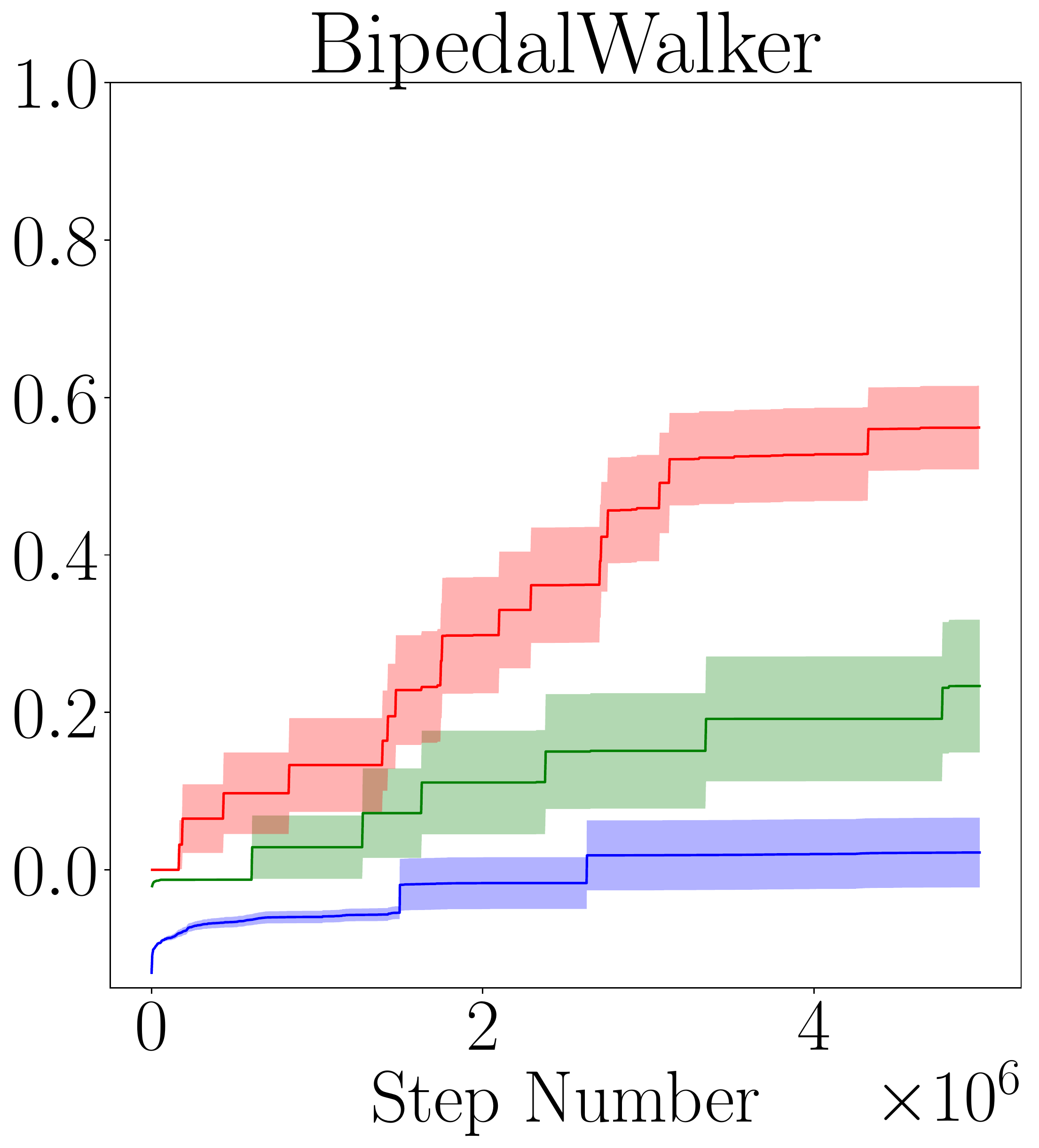}
    
    \includegraphics[width = .665\columnwidth]{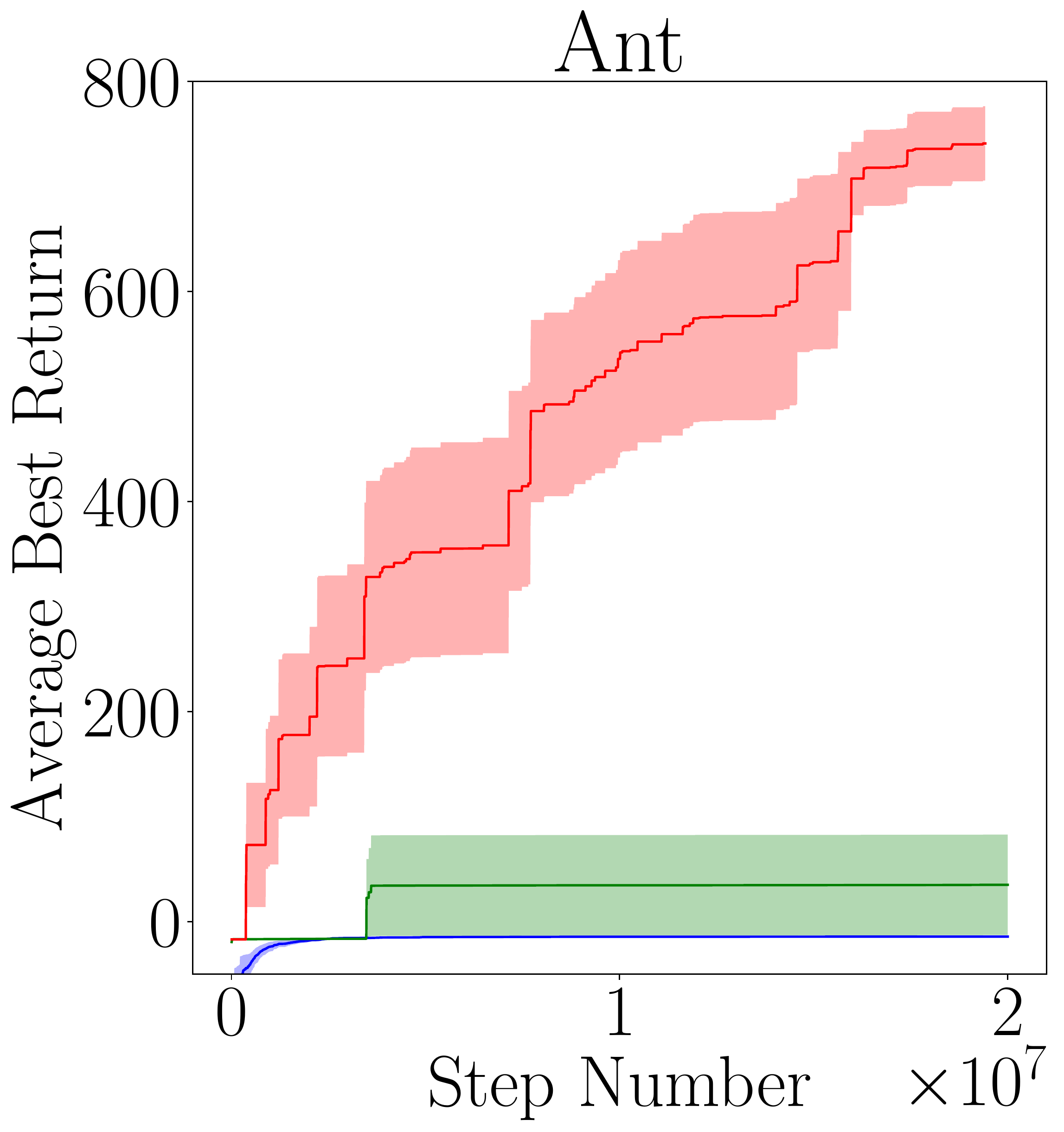}
    \includegraphics[width = .665\columnwidth]{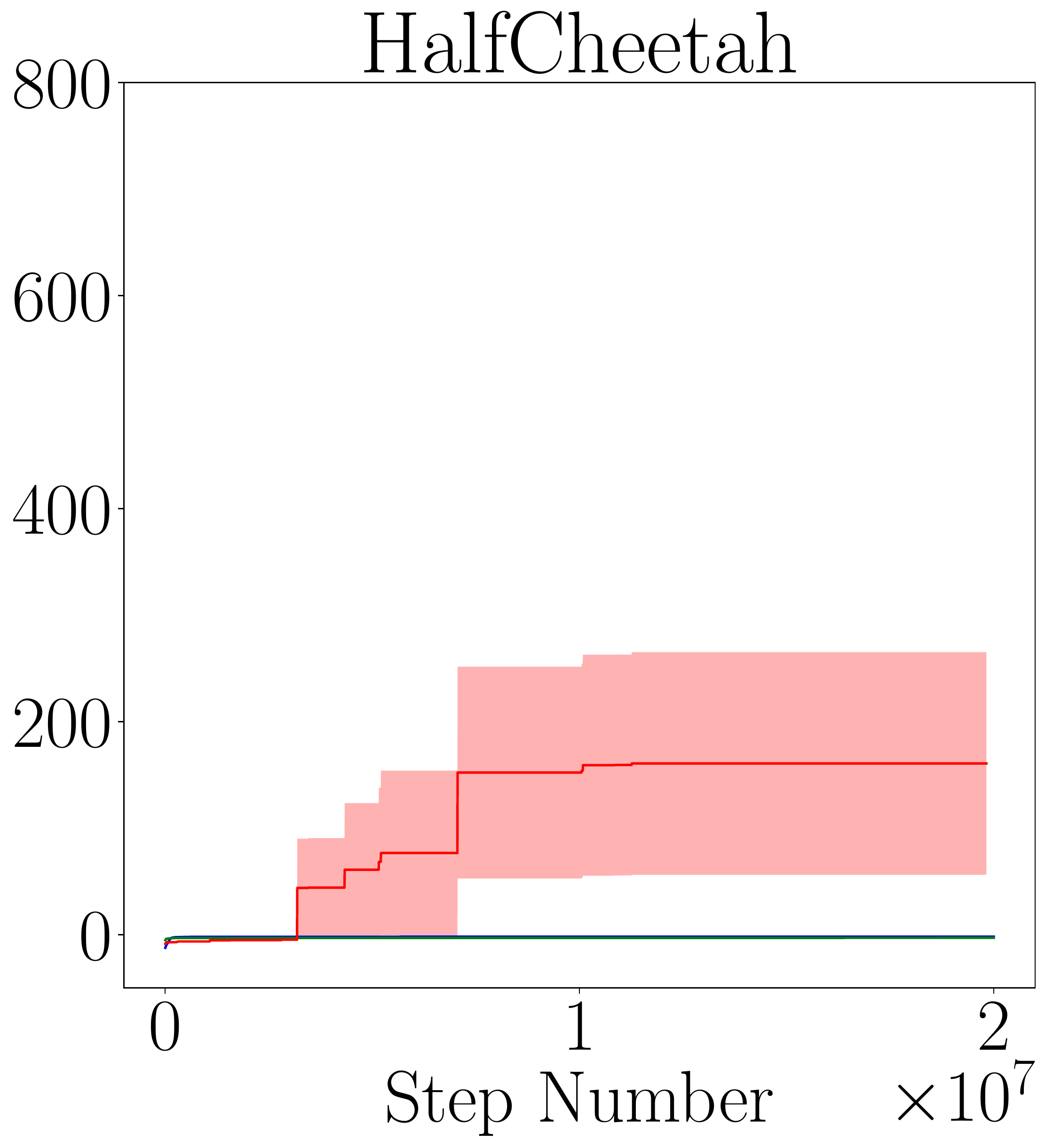}
    \includegraphics[width = .665\columnwidth]{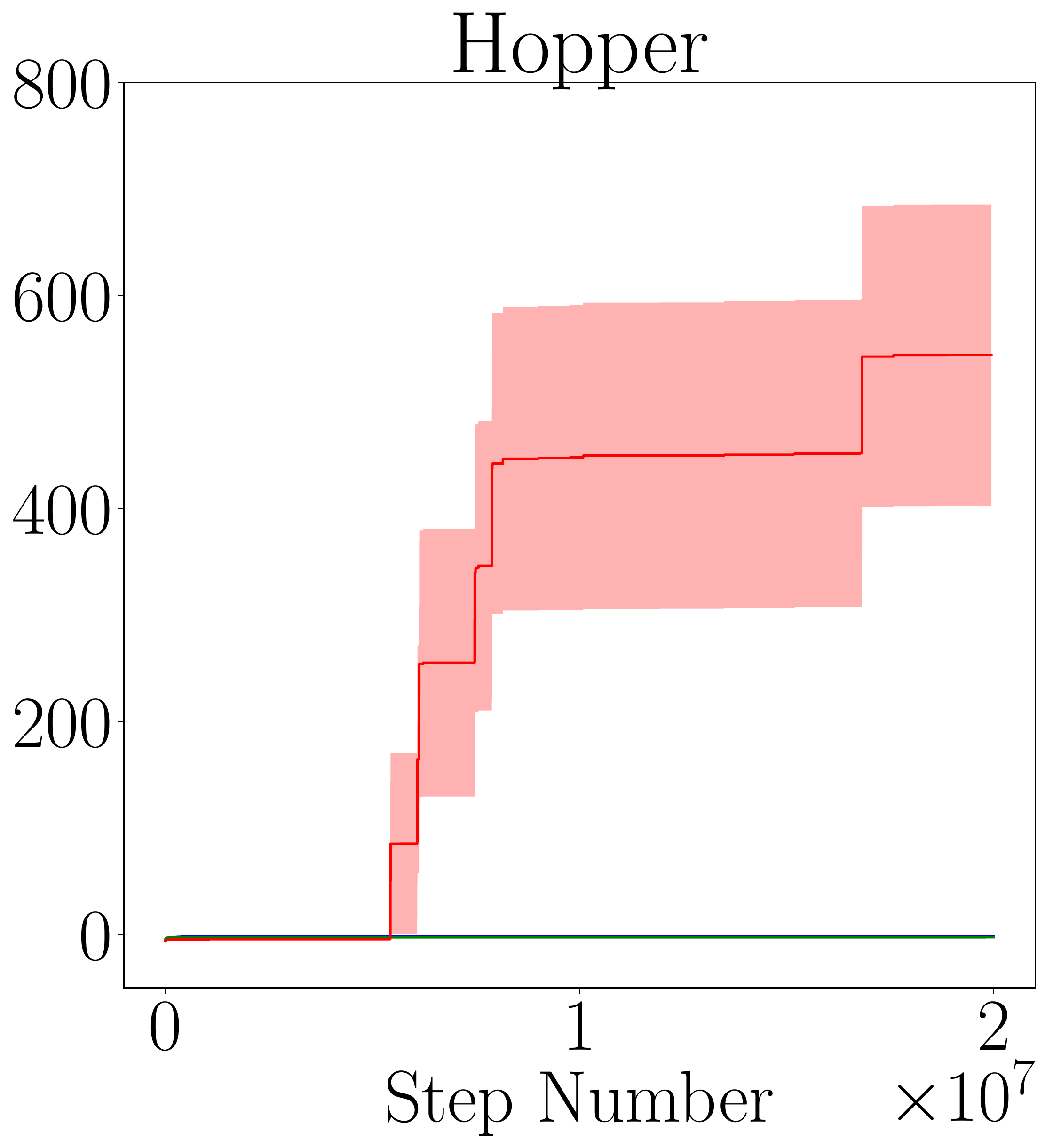}
    \caption{\small Performance curve for task environments using TRPO (blue), Deep GA (green), and MCTSPO (red) with best hyperparameters found. The hyperparemters for each algorithm are shown in \Cref{tab:results} in bold.}
    \label{fig:results}
\end{figure*}

Each classic control task is tested for 20 trials and each Roboschool task is tested for 10 trials, using different random seeds. 
Performance is evaluated using the average return of the best trajectory found. \Cref{tab:results} summarizes the results of different algorithms. 
To compare the performance and data efficiency of each algorithm, for each environment setting, we plot its training curve in \cref{fig:results}, showing the hyperparameter setting with the highest final return. 
The horizontal axis is the number of environment calls and the vertical axis is the average return of the best policy found so far. The shaded region of each curve is the error bound given by $\Delta/\sqrt{n_\text{trial}}$, where $\Delta$ is the standard deviation of the best returns.

\begin{table*}
\centering
\begin{tabular}{@{}llrrrrrr@{}}  
\toprule
Policy & Hyperparameter & Acrobot & MountainCar & BipedalWalker & Ant & HalfCheetah & Hopper\\
 \midrule
TRPO    & step size 0.1  &  \textbf{0.0}  & \textbf{0.0} & 0.0 & $-\textbf{14.3}$ & $-\textbf{1.90}$ & $-1.92$\\
        & step size 1.0   & 0.0 &  $-0.024$ & \textbf{0.022} & $-19.6$ & $-2.01$ & $-\textbf{1.48}$\\ \midrule
Deep GA & population 100  & 0.364 &  0.8 &  0.103 & \textbf{35.0} & $-3.69$ & $-2.76$\\
        & population 500  & 0.320 & 0.894 & \textbf{0.233} & $-14.1$ & $-3.12$ & $-2.54$\\
        & population 1000  & \textbf{0.593} & \textbf{0.942} & 0.153 & $-14.5$ & $-\textbf{2.98}$ & $-\textbf{2.46}$\\ 
 \midrule
MCTSPO  & $k=\alpha=0.3$  & 0.684 & 0.934 & \textbf{0.562} & \textbf{741} & \textbf{161} & \textbf{544}\\
        & $k=\alpha=0.5$  & \textbf{0.772} & \textbf{0.938} & 0.3 & $-16.5$ & $-3.37$ & 276\\
        & $k=\alpha=0.8$  & 0.638 & 0.937 & 0.042 & $-15.8$ & $-2.85$ & 91.1\\
\bottomrule
\end{tabular}
\caption{Average best return of TRPO, Deep GA, and MCTSPO using different hyperparameters on six task environments. The bold entries indicate the best performance found for each algorithm in each environment.\label{tab:results}}
\end{table*}

In the classic control environments, MCTSPO outperforms Deep GA and TRPO.
First, we note that TRPO almost gets stuck in local optima in all the experiment trials, which indicates the limitation of gradient based approaches on sparse and deceptive rewards. In Acrobot, MCTSPO is slightly better than Deep GA. In Mountain Car and Bipedal Walker, the advantage of MCTSPO is more evident. In Mountain Car, all MCTSPO trials converge to the optimal solution using approximately \num{4e6} environment calls fewer than Deep GA. In Bipedal Walker, MCTSPO achieves a much higher average return than Deep GA. In Roboschool, MCTSPO as has the best performance. It is able to reach the goal in most of the experiment trials except for the Half Cheetah, which is the hardest task attempted here. 

We observe that Deep GA also suffers from local optima. 
This sub-optimal behavior happens when the individuals chosen as parents are all in the ``valley" around the local optimum. 
Then, the descents are likely all distributed in the valley as well. 
Escaping the valley requires going against the direction of increasing return.
Even if some individuals in the next generation evolve in the right direction, they are not likely to be chosen as the parents. 
Thus, the entire population will be stuck in the local optimum. This situation is often caused by a poor initialization.

Novelty search is helpful in this case~\cite{GA}, where rewards are replaced by the novelty of the policy behavior, which encourages exploration. However, this method requires a behavior characteristic, which is domain-specific. 
In addition, Deep GA with novelty search essentially sweeps over the entire policy space, thus preventing exploitation or focusing search efforts on more promising regions. 

This problem is mitigated by MCTSPO. 
States or nodes with the same ancestor are in nearby regions of the policy space through constrained mutation. 
When a local optimum is found, the node around the local optimum is expanded more than other nodes. 
However, as more descendants are added to this node, the value estimation, $Q(s,a)$, associated with this node quickly reaches its limit, where $N(s,a)$ keeps increasing. As a result, the upper confidence bound estimation of $s$ decreases, which prevents more descendants being added to $s$ in the later iterations.
The algorithm then expands nodes in other regions. 
Since each mutation path is stored in the search tree, there is a non-negligible chance to find a long enough path that escapes the local optimum. 


As we decrease the difficulty of tasks, e.g. increase the power of the car in Mountain Car or decrease the target distance in Roboschool tests, the advantage of MCTSPO becomes less obvious. Actually when the environment goal could be easily achieved without much exploration, the sample efficiency of TRPO and Deep GA is higher than MCTSPO, although they converge to similar final performance. In this case, the exploration effort done by MCTSPO could be redundant. This could be mitigated by reducing the exploration constant as well as the progressive widening parameter in MCTSPO.

\section{Conclusion}
\label{sec:conclusion}
This paper discussed the limitations of popular gradient-based DRL solvers on environments with sparse or deceptive rewards.
We propose an approach called MCTSPO to balance exploration and exploitation of policy optimization using Monte-Carlo tree search in the policy space. 
We demonstrate how we can escape local optima and efficiently train policies in continuous control tasks.
The algorithm shows promising results in different classic control and robotics tasks, outperforming TRPO and Deep GA. 

In our implementation of MCTS, sampling is not executed in parallel, which limits the efficiency of MCTSPO as well as the training clock time.
Parallel MCTS techniques are likely to further improve the time efficiency~\cite{parallelMCTS}.
The combination of gradient-based methods 
with MCTSPO might further improve sample efficiency. For example, in the rollout, instead of \emph{no mutation}, we could run several iterations of gradient-based optimization to increase the accuracy of the value estimates.

\section*{Acknowledgments}
We thank anonymous reviewers for their helpful feedback and suggestions. This work is sponsored through the Stanford Center for AI Safety.
Zongzhang Zhang is in part supported by the National Natural Science Foundation of China under Grant No. 61876119, and the Natural Science Foundation of Jiangsu under Grant No. BK20181432, and the China Scholarship Council.


\bibliographystyle{named}
\bibliography{ms}

\begin{thebibliography}{}

\bibitem[\protect\citeauthoryear{Baier and Cowling}{2018}]{EMCTS}
Hendrik Baier and Peter~I. Cowling.
\newblock Evolutionary {MCTS} for multi-action adversarial games.
\newblock In {\em IEEE Conference on Computational Intelligence and Games
  (CIG)}, pages 1--8. IEEE, 2018.

\bibitem[\protect\citeauthoryear{Brockman \bgroup \em et al.\egroup
  }{2016}]{openaigym}
Greg Brockman, Vicki Cheung, Ludwig Pettersson, Jonas Schneider, John Schulman,
  Jie Tang, and Wojciech Zaremba.
\newblock Open{AI} gym.
\newblock {\em arXiv:1606.01540}, 2016.

\bibitem[\protect\citeauthoryear{Chaslot \bgroup \em et al.\egroup
  }{2008a}]{progressive_widening}
Guillaume~M.J.B Chaslot, Mark~H.M. Winands, H.~Jaap Van~Den Herick, Jos~W.H.M
  Uiterwijk, and Bruno Bouzy.
\newblock Progressive strategies for {M}onte-{C}arlo tree search.
\newblock {\em New Mathematics and Natural Computation}, 4(03):343--357, 2008.

\bibitem[\protect\citeauthoryear{Chaslot \bgroup \em et al.\egroup
  }{2008b}]{parallelMCTS}
Guillaume~M.J.B. Chaslot, Mark~H.M. Winands, and H.~Jaap Van Den~Herik.
\newblock Parallel {M}onte-{C}arlo tree search.
\newblock In {\em International Conference on Computers and Games}, pages
  60--71, 2008.

\bibitem[\protect\citeauthoryear{Chrabaszcz \bgroup \em et al.\egroup
  }{2018}]{es_benchmark}
Patryk Chrabaszcz, Ilya Loshchilov, and Frank Hutter.
\newblock Back to basics: {B}enchmarking canonical evolution strategies for
  playing {A}tari.
\newblock {\em arXiv:1802.08842}, 2018.

\bibitem[\protect\citeauthoryear{Cou{\"e}toux \bgroup \em et al.\egroup
  }{2011}]{MCTSdpw}
Adrien Cou{\"e}toux, Jean-Baptiste Hoock, Nataliya Sokolovska, Olivier Teytaud,
  and Nicolas Bonnard.
\newblock Continuous upper confidence trees.
\newblock In {\em Learning and Intelligent Optimization (LION)}, pages
  433--445, 2011.

\bibitem[\protect\citeauthoryear{Duan \bgroup \em et al.\egroup }{2016}]{rllab}
Yan Duan, Xi~Chen, Rein Houthooft, John Schulman, and Pieter Abbeel.
\newblock Benchmarking deep reinforcement learning for continuous control.
\newblock In {\em International Conference on Machine Learning (ICML)}, pages
  1329--1338, 2016.

\bibitem[\protect\citeauthoryear{Geramifard \bgroup \em et al.\egroup
  }{2015}]{Acrobot}
Alborz Geramifard, Christoph Dann, Robert~H. Klein, William Dabney, and
  Jonathan~P. How.
\newblock {RLP}y: {A} value-function-based reinforcement learning framework for
  education and research.
\newblock {\em Journal of Machine Learning Research}, 16:1573--1578, 2015.

\bibitem[\protect\citeauthoryear{Grzes and Kudenko}{2009}]{reward_shaping}
Marek Grzes and Daniel Kudenko.
\newblock Theoretical and empirical analysis of reward shaping in reinforcement
  learning.
\newblock In {\em International Conference on Machine Learning and
  Applications}, pages 337--344. IEEE, 2009.

\bibitem[\protect\citeauthoryear{Hessel \bgroup \em et al.\egroup
  }{2018}]{rainbow}
Matteo Hessel, Joseph Modayil, Hado Van~Hasselt, Tom Schaul, Georg Ostrovski,
  Will Dabney, Dan Horgan, Bilal Piot, Mohammad Azar, and David Silver.
\newblock Rainbow: Combining improvements in deep reinforcement learning.
\newblock In {\em AAAI Conference on Artificial Intelligence (AAAI)}, pages
  3215--3222, 2018.

\bibitem[\protect\citeauthoryear{Keller and Helmert}{2013}]{MaxUCT}
Thomas Keller and Malte Helmert.
\newblock Trial-based heuristic tree search for finite horizon {MDP}s.
\newblock In {\em International Conference on Automated Planning and Scheduling
  (ICAPS)}, pages 135--143, 2013.

\bibitem[\protect\citeauthoryear{Khadka and Tumer}{2018}]{ERL}
Shauharda Khadka and Kagan Tumer.
\newblock Evolution-guided policy gradient in reinforcement learning.
\newblock In {\em Advances in Neural Information Processing Systems (NIPS)},
  pages 1196--1208, 2018.

\bibitem[\protect\citeauthoryear{Kingma and Ba}{2015}]{adam}
Diederik~P. Kingma and Jimmy Ba.
\newblock Adam: A method for stochastic optimization.
\newblock In {\em International Conference on Learning Representations (ICLR)},
  2015.

\bibitem[\protect\citeauthoryear{Kochenderfer}{2015}]{MDP}
Mykel~J. Kochenderfer.
\newblock {\em Decision Making Under Uncertainty: Theory and Application}.
\newblock MIT Press, 2015.

\bibitem[\protect\citeauthoryear{Kocsis and Szepesv{\'a}ri}{2006}]{UCT}
Levente Kocsis and Csaba Szepesv{\'a}ri.
\newblock Bandit based {M}onte-{C}arlo planning.
\newblock In {\em European Conference on Machine Learning (ECML)}, pages
  282--293, 2006.

\bibitem[\protect\citeauthoryear{Lehman \bgroup \em et al.\egroup
  }{2018}]{safemutation}
Joel Lehman, Jay Chen, Jeff Clune, and Kenneth~O. Stanley.
\newblock Safe mutations for deep and recurrent neural networks through output
  gradients.
\newblock In {\em Genetic and Evolutionary Computation Conference}, pages
  117--124. ACM, 2018.

\bibitem[\protect\citeauthoryear{Mnih \bgroup \em et al.\egroup }{2015}]{DQN}
Volodymyr Mnih, Koray Kavukcuoglu, David Silver, Andrei~A. Rusu, Joel Veness,
  Marc~G. Bellemare, Alex Graves, Martin Riedmiller, Andreas~K. Fidjeland,
  Georg Ostrovski, et~al.
\newblock Human-level control through deep reinforcement learning.
\newblock {\em Nature}, 518(7540):529--533, 2015.

\bibitem[\protect\citeauthoryear{Moore}{1991}]{mountaincar}
Andrew Moore.
\newblock {\em Efficient Memory-based Learning for Robot Control}.
\newblock PhD thesis, Carnegie Mellon University, 1991.

\bibitem[\protect\citeauthoryear{Murthy and Raibert}{1984}]{Hopper}
Seshashayee~S. Murthy and Marc~H. Raibert.
\newblock 3{D} balance in legged locomotion: {M}odeling and simulation for the
  one-legged case.
\newblock {\em ACM SIGGRAPH Computer Graphics}, 18(1):27--27, 1984.

\bibitem[\protect\citeauthoryear{Salimans \bgroup \em et al.\egroup
  }{2017}]{ES}
Tim Salimans, Jonathan Ho, Xi~Chen, Szymon Sidor, and Ilya Sutskever.
\newblock Evolution strategies as a scalable alternative to reinforcement
  learning.
\newblock {\em arXiv:1703.03864}, 2017.

\bibitem[\protect\citeauthoryear{Schulman \bgroup \em et al.\egroup
  }{2015a}]{TRPO}
John Schulman, Sergey Levine, Pieter Abbeel, Michael Jordan, and Philipp
  Moritz.
\newblock Trust region policy optimization.
\newblock In {\em International Conference on Machine Learning (ICML)}, pages
  1889--1897, 2015.

\bibitem[\protect\citeauthoryear{Schulman \bgroup \em et al.\egroup
  }{2015b}]{Ant}
John Schulman, Philipp Moritz, Sergey Levine, Michael Jordan, and Pieter
  Abbeel.
\newblock High-dimensional continuous control using generalized advantage
  estimation.
\newblock {\em arXiv:1506.02438}, 2015.

\bibitem[\protect\citeauthoryear{Schulman \bgroup \em et al.\egroup
  }{2017}]{PPO}
John Schulman, Filip Wolski, Prafulla Dhariwal, Alec Radford, and Oleg Klimov.
\newblock Proximal policy optimization algorithms.
\newblock {\em arXiv:1707.06347}, 2017.

\bibitem[\protect\citeauthoryear{Such \bgroup \em et al.\egroup }{2017}]{GA}
Felipe~Petroski Such, Vashisht Madhavan, Edoardo Conti, Joel Lehman, Kenneth~O.
  Stanley, and Jeff Clune.
\newblock Deep neuroevolution: {G}enetic algorithms are a competitive
  alternative for training deep neural networks for reinforcement learning.
\newblock {\em arXiv:1712.06567}, 2017.

\bibitem[\protect\citeauthoryear{Wawrzynski}{2007}]{HalfCheetah}
Pawel Wawrzynski.
\newblock Learning to control a 6-degree-of-freedom walking robot.
\newblock In {\em EUROCON 2007-The International Conference on `` Computer as a
  Tool"}, pages 698--705. IEEE, 2007.

\bibitem[\protect\citeauthoryear{Wu \bgroup \em et al.\egroup }{2017}]{ACKTR}
Yuhuai Wu, Elman Mansimov, Roger~B. Grosse, Shun Liao, and Jimmy Ba.
\newblock Scalable trust-region method for deep reinforcement learning using
  {K}ronecker-factored approximation.
\newblock In {\em Advances in Neural Information Processing Systems (NIPS)},
  pages 5279--5288, 2017.

\end{thebibliography}
\end{document}